%% file: LH-ijcai21.tex
\documentclass{article}
\usepackage{ijcai21}

\usepackage{times}
\usepackage{soul}
\usepackage{url}
\usepackage[hidelinks]{hyperref}
\usepackage[utf8]{inputenc}
\usepackage[small]{caption}
\usepackage{graphicx}
\usepackage{amsmath}
\usepackage{amsthm}
\usepackage{booktabs}
\usepackage{algorithm}
\usepackage{algorithmic}
\input{dlab_macros}

\usepackage{subcaption}
\usepackage{amssymb}
\usepackage{bm}
\usepackage{bbm}
\usepackage{amsfonts}       
\usepackage{nicefrac}       
\usepackage{microtype}      
\usepackage{hyphenat}
\usepackage{latexsym}
\usepackage{mathptmx}
\usepackage{mathtools}
\usepackage{thmtools} 
\usepackage{thm-restate}

\newcommand{\newcite}[1]{\citeauthor{#1} \shortcite{#1}}

\newtheorem*{theorem*}{Theorem}

\usepackage{scalefnt}
\newcommand\smaller[2][0.8]{{\scalefont{#1}#2}}

\newcommand{\BERT}{\cpt{BERT}\xspace}
\newcommand{\BERTSS}{\cpt{BERT}\hyp{}\smaller{1}\cpt{S}\xspace}
\newcommand{\BERTPS}{\cpt{BERT\hyp{}PS}\xspace}
\newcommand{\BOW}{\cpt{BOW}\xspace}
\newcommand{\LSTM}{\cpt{LSTM}\xspace}
\newcommand{\GPT}{\cpt{GPT}\smaller{2}\xspace}
\newcommand{\GPTSS}{\cpt{GPT}\smaller{2}\hyp{}\smaller{1}\cpt{S}\xspace}
\newcommand{\GPTPS}{\cpt{GPT}\smaller{2}\hyp{}\cpt{PS}\xspace}
\newcommand{\dBERT}{distil\cpt{BERT}\xspace}
\newcommand{\Transformer}{\cpt{Transformer}\xspace}
\newcommand{\ROBERTA}{\cpt{R}o\cpt{BERT}a\xspace}

\title{Laughing Heads: Can Transformers Detect What Makes a Sentence Funny?}

\author{
Maxime Peyrard{\normalfont ,}
Beatriz Borges{\normalfont ,}
Kristina Gligorić {\normalfont and}
Robert West
\affiliations
EPFL
\emails
\{maxime.peyrard, beatriz.borges, kristina.gligoric, robert.west\}@epfl.ch
}

\begin{document}

\maketitle

\begin{abstract}
The automatic detection of humor poses a grand challenge for natural language processing.
Transformer\hyp based systems have recently achieved remarkable results on this task, but they usually
(1)~were evaluated in setups where serious \vs\ humorous texts came from entirely different sources, and
(2)~focused on benchmarking performance without providing insights into how the models work.
We make progress in both respects by training and analyzing transformer\hyp based humor recognition models on a recently introduced dataset consisting of minimal pairs of aligned sentences, one serious, the other humorous.
We find that, although our aligned dataset is much harder than previous datasets, transformer\hyp based models recognize the humorous sentence in an aligned pair with high accuracy (78\%).
In a careful error analysis, we characterize easy \vs\ hard instances.
Finally, by analyzing attention weights, we obtain important insights into the mechanisms by which transformers recognize humor.
Most remarkably, we find clear evidence that one single attention head learns to recognize the words that make a test sentence humorous, even without access to this information at training time. 
\end{abstract}

\input{010introduction}
\input{020relwork}
\input{030data}

\input{040models}
\input{050experiments}
\input{060conclusion}

\vfill
\section*{Acknowledgments}
Thanks to Zachary Horvitz and Eric Horvitz for insightful feedback.
With support from
Swiss National Science Foundation (grant 200021\_185043),
European Union (TAILOR, grant 952215),
and gifts from
Google, Facebook, Microsoft

\clearpage

{\small
\bibliographystyle{named}
\bibliography{LH}
}

\clearpage

\appendix
\input{100appendix}

\end{document}

%% file: dlab_macros.tex
\usepackage[utf8]{inputenc}
\usepackage[T1]{fontenc}
\usepackage{hyphenat}
\usepackage{xspace}
\usepackage{amsfonts}
\usepackage{hyperref}
\usepackage{booktabs}
\usepackage{multirow}
\usepackage{makecell}
\usepackage{caption}
\usepackage{minibox}
\usepackage{bbm}
\usepackage{graphicx}
\usepackage{balance}
\usepackage{mathtools}
\usepackage{color}
\usepackage{marvosym}
\usepackage{ifthen}
\usepackage{textcomp}
\usepackage{enumitem}
\usepackage{verbatim}
\usepackage{algorithm}
\usepackage{algorithmic}
\usepackage{numprint}
\usepackage{balance}

\usepackage{amsthm}
\theoremstyle{plain}

\newcommand{\chatoDisplayMode}[1]{#1}

\definecolor{MyRed}{rgb}{0.6,0.0,0.0} 
\definecolor{MyBlack}{rgb}{0.1,0.1,0.1} 
\newcommand{\inred}[1]{{\color{MyRed}\sf\textbf{\textsc{#1}}}}
\newcommand{\frameit}[2]{
  \begin{center}
  {\color{MyRed}
  \framebox[.9\columnwidth][l]{
    \begin{minipage}{.85\columnwidth}
    \inred{#1}: {\sf\color{MyBlack}#2}
    \end{minipage}
  }\\
  }
  \end{center}
}

\newcommand{\note}[2][]{\chatoDisplayMode{\def\@tmpsig{#1}\frameit{{\Pointinghand} Note}{#2\ifx \@tmpsig \@empty \else \mbox{ --\em #1}\fi}}}
\newcommand{\todo}[2][]{\chatoDisplayMode{\def\@tmpsig{#1}\frameit{{\Writinghand} To-do}{#2\ifx \@tmpsig \@empty \else \mbox{ --\em #1}\fi}}}

\newcommand{\abbrevStyle}[1]{#1}

\newcommand{\ie}{\abbrevStyle{i.e.}\xspace}
\newcommand{\eg}{\abbrevStyle{e.g.}\xspace}
\newcommand{\cf}{\abbrevStyle{cf.}\xspace}

\newcommand{\vs}{\abbrevStyle{vs.}\xspace}

\newcommand{\Secref}[1]{Sec.~\ref{#1}}

\newcommand{\Tabref}[1]{Table~\ref{#1}}
\newcommand{\Figref}[1]{Fig.~\ref{#1}}

\newcommand{\Appref}[1]{Appendix~\ref{#1}}

\newcommand{\xhdr}[1]{\vspace{1.7mm}\noindent{{\bf #1.}}}

\newcommand{\denselist}{ \itemsep -2pt\topsep-10pt\partopsep-10pt }

\newcommand{\textcite}[1]{\citeauthor{#1} \shortcite{#1}}

\newcommand{\cpt}[1]{\textsc{\MakeLowercase{#1}}}

\newcommand{\hide}[1]{}

\makeatletter
\newcommand{\iffont}[2]{\ifthenelse{\equal{\f@family}{#1}}{#2}{}}
\makeatother

  \usepackage{mathptmx}

  \DeclareSymbolFont{greek}{OML}{cmm}{m}{n}
  \DeclareMathSymbol{\alpha}{\mathalpha}{greek}{"0B}
  \DeclareMathSymbol{\beta}{\mathalpha}{greek}{"0C}
  \DeclareMathSymbol{\gamma}{\mathalpha}{greek}{"0D}
  \DeclareMathSymbol{\delta}{\mathalpha}{greek}{"0E}
  \DeclareMathSymbol{\epsilon}{\mathalpha}{greek}{"0F}
  \DeclareMathSymbol{\zeta}{\mathalpha}{greek}{"10}
  \DeclareMathSymbol{\eta}{\mathalpha}{greek}{"11}
  \DeclareMathSymbol{\theta}{\mathalpha}{greek}{"12}
  \DeclareMathSymbol{\iota}{\mathalpha}{greek}{"13}
  \DeclareMathSymbol{\kappa}{\mathalpha}{greek}{"14}
  \DeclareMathSymbol{\lambda}{\mathalpha}{greek}{"15}
  \DeclareMathSymbol{\mu}{\mathalpha}{greek}{"16}
  \DeclareMathSymbol{\nu}{\mathalpha}{greek}{"17}
  \DeclareMathSymbol{\xi}{\mathalpha}{greek}{"18}
  \DeclareMathSymbol{\pi}{\mathalpha}{greek}{"19}
  \DeclareMathSymbol{\rho}{\mathalpha}{greek}{"1A}
  \DeclareMathSymbol{\sigma}{\mathalpha}{greek}{"1B}
  \DeclareMathSymbol{\tau}{\mathalpha}{greek}{"1C}
  \DeclareMathSymbol{\upsilon}{\mathalpha}{greek}{"1D}
  \DeclareMathSymbol{\phi}{\mathalpha}{greek}{"1E}
  \DeclareMathSymbol{\chi}{\mathalpha}{greek}{"1F}
  \DeclareMathSymbol{\psi}{\mathalpha}{greek}{"20}
  \DeclareMathSymbol{\omega}{\mathalpha}{greek}{"21}
  \DeclareMathSymbol{\varepsilon}{\mathalpha}{greek}{"22}
  \DeclareMathSymbol{\vartheta}{\mathalpha}{greek}{"23}
  \DeclareMathSymbol{\varpi}{\mathalpha}{greek}{"24}
  \DeclareMathSymbol{\varrho}{\mathalpha}{greek}{"25}
  \DeclareMathSymbol{\varsigma}{\mathalpha}{greek}{"26}
  \DeclareMathSymbol{\varphi}{\mathalpha}{greek}{"27}
  \DeclareSymbolFont{otone}{OT1}{cmr}{m}{n}
  \DeclareMathSymbol{\Gamma}{\mathalpha}{otone}{0}
  \DeclareMathSymbol{\Delta}{\mathalpha}{otone}{1}
  \DeclareMathSymbol{\Theta}{\mathalpha}{otone}{2}
  \DeclareMathSymbol{\Lambda}{\mathalpha}{otone}{3}
  \DeclareMathSymbol{\Xi}{\mathalpha}{otone}{4}
  \DeclareMathSymbol{\Pi}{\mathalpha}{otone}{5}
  \DeclareMathSymbol{\Sigma}{\mathalpha}{otone}{6}
  \DeclareMathSymbol{\Upsilon}{\mathalpha}{otone}{7}
  \DeclareMathSymbol{\Phi}{\mathalpha}{otone}{8}
  \DeclareMathSymbol{\Psi}{\mathalpha}{otone}{9}
  \DeclareMathSymbol{\Omega}{\mathalpha}{otone}{10}
  \DeclareSymbolFont{syms}{OML}{cmm}{m}{it}
  \DeclareMathSymbol{\partial}{\mathord}{syms}{"40}
  \DeclareMathAlphabet{\mathbold}{OML}{cmm}{b}{it}
  \DeclareSymbolFont{largesymbols}{OMX}{cmex}{m}{n}


%% file: 010introduction.tex
\section{Introduction}
\label{sec:Introduction}

Humor is a unique feature of human cognition and communication. 
The computational aspects of humor form a promising research area to advance the semantic understanding, common-sense reasoning, and language abilities of artifical intelligence (AI) systems. 
In particular, computational tasks such as humor detection and generation offer a rich and challenging testing ground for modern language understanding systems, to the extent that
the ability to understand humor is commonly believed to be ``AI-complete'' \cite{getting_serious}.

Humorous texts often display complex narrative structures, rely on shared common-sense knowledge, and exploit subtle lexico\hyp grammatical features of language \cite{ThePrimerofHumorResearch}.
Nonetheless, impressive progress in humor detection has recently been made thanks to the advent of transformer architectures \cite{NIPS2017_7181} and the pretraining--finetuning paradigm \cite{devlin-etal-2019-bert}. For instance, \newcite{weller-seppi-2019-humor} finetuned a transformer to classify short texts as funny or serious, obtaining high accuracy on a dataset of Reddit jokes. They, however, neither performed error analysis nor gave insights into what signals are discovered and exploited by their model. 
In order to guide future research, a better understanding of how current models behave in practice, as well as a sharper outline of their capabilities, appears necessary. 


In this work, we propose to leverage aligned pairs of funny and serious sentences collected via Unfun.me, an online game where players make minimal edits to satirical news headlines with the goal of making other players believe the results are serious headlines \cite{west2019reverse}. Previously, \newcite{west2019reverse} released and analyzed 2.8K pairs from Unfun.me in order to better understand how humans create humor. Here, we use an extended dataset with more than 20K new pairs to probe the capacities of modern transformers such as \BERT \cite{devlin-etal-2019-bert}. The minimal modifications between funny and serious headlines allow us to zoom in precisely on where the humor happens and thus perform entirely new kinds of analysis. 

\newcite{west2019reverse} found that the difference between funny and serious headlines generated by humans tend to be explained by an opposition along certain dimensions important to the human condition, e.g., \emph{reasonable vs.\ absurd} or \emph{non-obscene vs.\ obscene}. We use the annotated, aligned pairs and the identified dimensions of opposition in order to perform fine-grained error analysis of transformer models.

\xhdr{Contributions}
We finetune and evaluate standard \BERT variants for two different humor detection tasks: (i) the \emph{single\hyp sentence setup}, where models detect whether an input sentence is funny or not, and (ii) the \emph{paired setup} (cf.\ \Secref{sec:models}), where models detect for an input pair of aligned funny and serious sentences which one is funny.  
Based on the natural pairing of the data, we obtain diverse insights about the data and models:

    
    
    
    
    

\begin{enumerate}
    \setlength\itemsep{-0.1em}
    \item In a careful error analysis, we characterize easy vs.\ hard instances (\Secref{ssec:perf_per_type} and \ref{ssec:perf_vs_size}).
    \item Inspection of attention heads reveals that the critical computation takes place in the last transformer layers, confirming that the model picks up on semantic, rather than lexical or syntactic, features (\Secref{ssec:attention_pattern}).
    \item A particularly striking result is the emergence, during finetuning, of a head (the ``laughing head'') specialized in attending to the funny part of an input sentence. This head alone detects the humorous part of a sentence five times more accurately than a random baseline (\Secref{ssec:special_head}).
\end{enumerate}


Code and data,\footnote{ \url{https://github.com/epfl-dlab/laughing-head}}
as well as an extended version of the paper (with the appendices referenced here),%
\footnote{
\label{fn:app}
\url{https://arxiv.org/abs/2105.09142}}
are available online.

%% file: 020relwork.tex
\section{Related Work}
\label{sec:related_work}

We discuss two aspects of previous research efforts on computational humor relevant to our work: datasets and approaches. 

\xhdr{Humor detection datasets}
Dataset creation approaches usually extract texts considered humorous online, e.g., on Twitter \cite{potash-etal-2017-semeval,zhang_twitter} or Reddit \cite{weller-seppi-2019-humor}, and, in parallel, collect serious texts from other sources to induce balanced datasets. For instance, \newcite{mihalcea2005making} collected 16K one-liner punchlines and matched them with 16K news headlines. Similarly, \newcite{yang-etal-2015-humor} matched 2.4K puns with 2.4K headlines. Finally, the ``Stierlitz'' dataset \cite{blinov-etal-2019-large} consists of 60K Russian jokes paired with news headlines.
Others have directly collected pairs of funny and serious sentences. \newcite{west2019reverse} collected pairs via Unfun.me, an online game where players propose small edits to turn satirical news headlines into serious-looking ones.
The Humicroedit \cite{hossain-etal-2019-president}
and FunLines \cite{hossain2020stimulating}
datasets were
obtained via the reverse approach, where humans edited serious headlines into funny ones. The researchers have used their datasets to study humor perception and creation. 
Here, we use an updated version of the data collected by \newcite{west2019reverse} to study automatic humor detection methods.

\xhdr{Humor detection approaches}
Researchers have applied humor detection techniques to several kinds of humor, e.g.,  irony \cite{wallace-etal-2015-sparse,reyes2013}, sarcasm \cite{gonzalez-ibanez-etal-2011-identifying},
 or satire \cite{goldwasser-zhang-2016-understanding}.
Various baselines have been proposed, such as statistical $n$-gram analysis \cite{Taylor04computationallyrecognizing} or classifiers based on human-crafted features \cite{purandare-litman-2006-humor,kiddon-brun-2011-thats}. Following the evolution of the NLP field, pretrained word vectors lead to further improvements \cite{yang-etal-2015-humor,cattle-ma-2018-recognizing}. Finally, deep learning approaches have become ubiquitous \cite{chen-soo-2018-humor,blinov-etal-2019-large}. In particular, transformers \cite{NIPS2017_7181} such as \BERT \cite{devlin-etal-2019-bert} are now used to recognize funny sentences \cite{weller-seppi-2019-humor} and generate funny texts \cite{horvitz-etal-2020-context}
Many researchers have aimed to understand how humans generate and perceive humor \cite{ThePrimerofHumorResearch}. Some of the resulting theories have been empirically verified. For instance, an analysis of the Unfun.me dataset~\cite{west2019reverse} produced evidence in favor of the \emph{General Theory of Verbal Humor} \cite{attardo1991script}. The original and modified (``unfunned'') headlines are generally opposed
to each other along certain dimensions important to
the human condition (e.g., \emph{reasonable vs.\ absurd}, \emph{high vs.\ low stature}, or \emph{non-obscene vs.\ obscene}).
On the contrary, despite impressive recent progress on automatic humor detection, the trained models have not been analyzed in depth. For instance, \newcite{weller-seppi-2019-humor} and \newcite{blinov-etal-2019-large} focus on reporting the performance of their transformer models. Yet, to further advance the field of computational humor and NLP in general, it is important to understand the capabilities of these models. 

%% file: 030data.tex
\section{Data}
\label{sec:data}
To investigate humor detection using transformer models, we employ data particularly well-suited for a fine-grained understanding in controlled settings: pairs of funny and serious sentences with a small lexical difference collected via the Unfun.me game and previously released by \newcite{west2019reverse}. Unfun.me is an online game that incentivizes players to make minimal edits to satirical news headlines with the goal of making other players believe the results are serious news headlines. 

We use an extended dataset of 23,113 pairs \cite{unfun_reference},%
which we randomly split into 18,832 training pairs, 2,414 validation pairs, and 1,867 testing pairs. Additionally, as part of the game, a subset of pairs has been annotated with quality ratings measuring how well the unfunning process worked, i.e., whether the unfunned sentence was perceived as serious by other humans. These annotations come from other players who evaluated the quality of the unfunned sentences (for details, see \newcite{west2019reverse}). From these annotations, we form a restrictive \emph{high-quality test set} of instances that received the maximum score according to all annotators, consisting of $754$ pairs. We later refer to this test set as ``HQ''.

Finally, a subset of the test set (254 pairs) comes with manual annotations from two trained annotators capturing the opposition that leads to humor in the pair (\cf\ \Secref{sec:Introduction}). In this work, we use the following seven types of opposition (listed with examples; satirical versions in bold; multiple oppositions may apply to the same pair; ``$\emptyset$'' refers to empty string):
\begin{enumerate}
    \setlength\itemsep{0.0em}
    \item normal/\textbf{abnormal}: \emph{Bush picks} \{\emph{\textbf{laser}, rural}\} \emph{background for presidential portrait}
    \item possible/\textbf{impossible}: \emph{City opens new art} \{\emph{\textbf{jail}, museum}\}
    \item non-violence/\textbf{violence}: \emph{Russian officials promise low} \{\emph{\textbf{death}, highway}\} \emph{toll for Olympics}
    \item good/\textbf{bad} intentions: \emph{BP ready to resume oil} \{\emph{\textbf{spilling}, drilling}\}
    \item reasonable/\textbf{absurd} response: \emph{general motors reports record sales of new} \{\emph{\textbf{disposable}, $\emptyset$}\} \emph{car}
    \item high/\textbf{low} stature: \emph{Hollywood mourns passing of} \{\emph{\textbf{16th or 17th Lassie}, Robin Williams}\}
    \item non-obscene/\textbf{obscene}: \emph{Tiger Woods announces return to} \{\emph{\textbf{sex}, golf}\}
\end{enumerate}

The advantage of such data for our analysis are threefold: (1) a naturally paired setup (2) with minimal modifications between funny and serious sentences and (3) additional annotations allowing us to investigate the performance of machine learning systems in fine-grained ways.

%% file: 040models.tex
\section{Models}
\label{sec:models}

We train standard transformer models to detect humor in the two setups described below.

\xhdr{Single\hyp sentence setup}
We first ignore the pairing between funny and serious sentences. The sentences are only associated with a binary label indicating whether they are funny or not. An encoder $E$ maps a sentence $s_i$ into a feature space, and a classifier converts the sentence representation $E(s_i)$ into a binary prediction. We train the model with the binary cross\hyp entropy loss by finetuning the full pretrained model with backpropagation. 


\xhdr{Paired setup}
Next, we exploit the natural pairing of our data. Here, each funny--serious pair is first randomly ordered and then associated with a binary label indicating whether the first sentence is the funny one.
This setup is naturally modeled via Siamese networks. For a given encoder $E$ and an input pair $(s_i, s_j)$, a classifier is trained on top of the concatenation of the feature representation of the two sentences, $[E(s_i), E(s_j)]$. Siamese networks have been successfully used recently in tasks based on the comparison of two sentences \cite{reimers-gurevych-2019-sentence}.


\xhdr{Encoders used}
The main focus of our analysis is on the \BERT model, thus we employ \BERT-base, an encoder with 12 layers and 12 heads per layer. Additionally, we report the performances of two \BERT variants: \dBERT, a simplified and smaller alternative to \BERT, and \ROBERTA, a variant of \BERT pretrained with an improved training setup.
Each of these can be either used in the single\hyp sentence setup (denoted ``1S'') or in the paired setup (denoted ``PS''). For each setup, the \BERT variants can be either fully finetuned or kept with weights frozen and only the classifier layer being finetuned.
For reference, we also report the performance of baseline encoders without pretraining: \BOW represents sentences by the average of their fastText word vectors \cite{grave2018learning}; \LSTM is a vanilla LSTM architecture trained on top of pretrained fastText vectors; and \Transformer is a vanilla transformer with the same architecture as \BERT, but trained from scratch without pretraining.
Language models such as \GPT \cite{radford2019language} are also important baselines, as it is often believed that humorous sentences are more surprising (\ie, have a lower probability according to the language model). In the paired setup, a natural baseline thus predicts the sentence with the lower \GPT likelihood to be funny. In the single\hyp sentence setup, a simple approach consists of predicting funny whenever the sentence probability is below some threshold previously identified via search on the training set. Later, we also use sentence probability scores provided by \GPT in the analysis (cf.\ \Secref{sec:analysis}).

\xhdr{Error bars and significance level}
Throughout the rest of the paper, error bars in plots represent bootstrapped $99\%$ confidence intervals, and when we say that a change or difference is ``significant'', we technically mean that $p<0.01$ in a paired $t$-test for comparing means.



\begin{table}[t]
\centering
\setlength{\tabcolsep}{10pt}
\begin{tabular}{@{}lcc|cc@{}}
\toprule
    & \multicolumn{2}{c}{1S} & \multicolumn{2}{c}{PS} \\
    & Full & HQ & Full & HQ \\
\midrule                            
\multicolumn{3}{l}{\emph{\textbf{No pretraining}}} \\ 
\hspace{6mm} \BOW                   &  .511 & .509     & .515 & .513 \\
\hspace{6mm} \LSTM                  &  .512 & .511     & .606 & .598 \\
\hspace{6mm} \Transformer           &  \textbf{.522} & \textbf{.526}     & \textbf{.611} & \textbf{.607} \\
\midrule                            
\multicolumn{3}{l}{\emph{\textbf{Pretraining, no finetuning}}} \\
\hspace{6mm} \GPT                   &  .526 & .522      & \textbf{.704} & \textbf{.682} \\
\hspace{6mm} \BERT                  &  .536 & .531      & .689 & .675 \\
\hspace{6mm} \dBERT                 &  .534 & .529      & .685 & .669  \\
\hspace{6mm} \ROBERTA               &  \textbf{.575} & \textbf{.568}      & .684 & .675  \\
\midrule
\multicolumn{3}{l}{\emph{\textbf{Pretraining and finetuning}}} \\
\hspace{6mm} \BERT                 &  .645 & .641      & .766 & .737 \\
\hspace{6mm} \dBERT                &  \textbf{.651} & \textbf{.647}      & \textbf{.777} & \textbf{.758}  \\
\hspace{6mm} \ROBERTA              &  .649 & .640      & .755 & .751  \\

\bottomrule                            
\end{tabular}
\caption{Accuracy of various standard models for both setups: single\hyp sentence (1S) and paired (PS). Datasets are balanced, so random baselines have accuracy $0.5$. Best values per block in bold.}
\label{tab:eval-systems}
\end{table}

\subsection{Accuracy on Unfun.me Dataset}
\label{ssec:general_perf}
In \Tabref{tab:eval-systems}, we report the accuracy of each encoder described above in both the single\hyp sentence and the paired setup for both the full test set (\textit{Full}) and the subset of the test set whose pairs were annotated as high\hyp quality (\textit{HQ}).

The results indicate that humor detection is a challenging task, especially in the single\hyp sentence setup: simple baselines such as \BOW, \LSTM, and \Transformer barely improve upon random prediction. Only finetuned \BERT architectures are significantly better than random, with the exception of \ROBERTA, which achieves above\hyp chance accuracy (57.5\%) even without finetuning.
In comparison, systems perform much better in the paired setup, where the classifier can take its decision based on the information from both sentences. The best model, \dBERT with finetuning, achieves an accuracy of 77.7\%. Both \BERT and \ROBERTA achieve similar accuracy, with no significant difference.
Overall, performance on the full test set and the HQ subset is similar, with a tendency towards slightly lower performance on the high\hyp quality instances. 
However, the differences between the high\hyp quality set and the full set are not significant, indicating that the full test set performance is a good indicator of performance on high-quality instances. 
Given the popularity of \BERT and the similar performance of \BERT, \dBERT, and \ROBERTA, we focus on \BERT encoders in the rest of the paper:
\BERT with no finetuning (simply called \BERT), \BERT finetuned on the single\hyp sentence setup (\BERTSS), and \BERT finetuned on the paired setup (\BERTPS). 

\begin{table}
        \small
        \centering
        \begin{tabular}{l|cc|cc}
        \toprule
        &  \multicolumn{2}{c|}{1S} & \multicolumn{2}{c}{PS} \\
         Type & \GPT & \cpt{BERT} & \GPT & \cpt{BERT} \\
        \midrule
        normal/abnormal & .493 & .630 & .815 & .849 \\
        possible/impossible & .518 & .665 & .790 & .821 \\
        non-violence/violence & \textbf{.553} & .657 & .842 & .816 \\
        good/bad intentions & .532 & .606 & .702 & .723 \\
        reasonable/absurd & .537 & \textbf{.704} & .889 & .907 \\
        high/low stature & .510 & .647 & .872 & .892 \\
        non-obscene/obscene & .516 & .582 & \textbf{.918} & \textbf{.989} \\
        \bottomrule
        \end{tabular}
        \caption{Accuracy per humor type as annotated in the Unfun.me dataset, for both the single\hyp sentence (1S) and paired (PS) setup, and for finetuned \BERT encoders and \GPT-based baselines.}\label{tab:error-analysis}
\end{table}

\subsection{Accuracy by Humor Type}
\label{ssec:perf_per_type}
We report performance per humor type in 
\Tabref{tab:error-analysis}, which compares the performance of \BERTSS and \BERTPS against the \GPT baselines (\GPTSS predicts funny if the sentence is less likely than a threshold chosen to maximize accuracy on the training data; \GPTPS predicts the less likely sentence in a pair to be the funny one).
In general, performance is higher than on the full test set, probably because these are more standard instances of humor likely to be seen often in the training set.
We also observe that different models perform best for different types. For instance, in the paired setup, models perform well when detecting humor in \textit{non-obscene\slash obscene} pairs, but this type is one of the hardest in the single\hyp sentence setup.
Interestingly, the \GPTPS baseline works better than \BERTPS for the \textit{non-violence\slash violence} type of humor, which might be explained by the prevalence of violence in general text, sometimes used for funny purposes, and sometimes not, such that the model has difficulty capturing violence as a dimension of humor.
Also, there is little improvement from \GPTPS to \BERTPS for the \textit{reasonable\slash absurd} type of humor, possibly because this type is mostly marked by sentence surprisal, which is explicitly captured by \GPT. 
Finally, the best model, \BERTPS, performs significantly worse for the more abstract \textit{good\slash bad intentions} type than for other types. 

These findings highlight the particular strengths and weaknesses of existing models and can inform future work. We release a simple tool to repeat this analysis, so other researchers can easily benchmark their new models of humor detection.


%% file: 050experiments.tex
\section{Analysis of Transformer Model Behavior}
In this section, we leverage the structure of the Unfun.me data to perform a deeper analysis of \BERTSS and \BERTPS. 

\label{sec:analysis}
\begin{figure}
    \centering
    \includegraphics[width=0.55\linewidth]{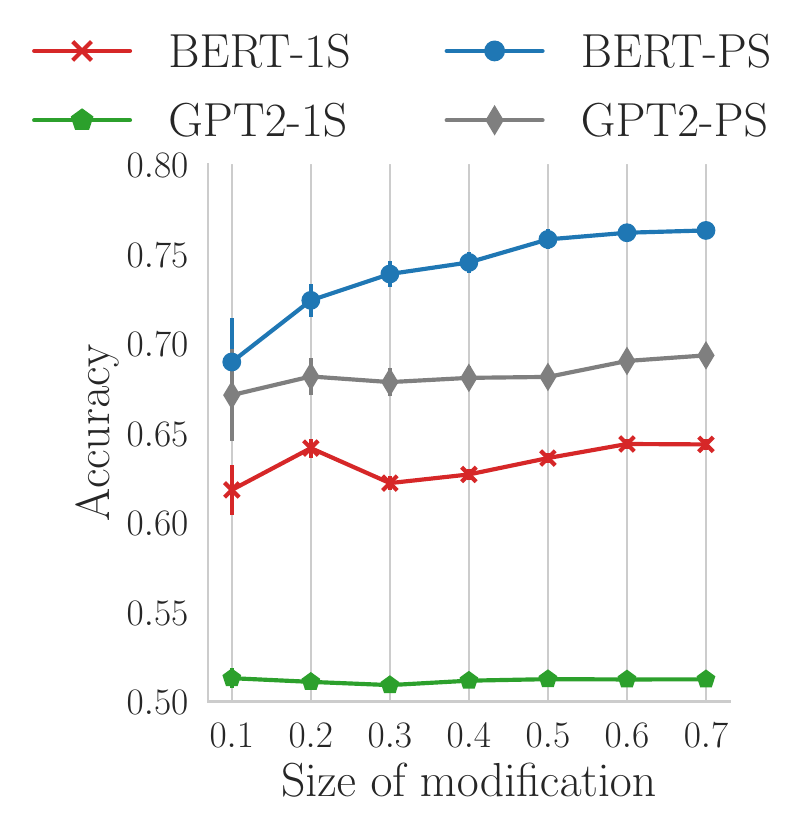}
    \caption{Humor detection accuracy as function of size of modification (lexical difference measured as Jaccard distance between token sets of funny and serious sentences in pairs).}
    \label{fig:perf_mod_size}
\end{figure}



\subsection{Models Perform Better for Pairs with Large Modifications}
\label{ssec:perf_vs_size}
We begin by measuring whether finetuned models perform differently for pairs with small vs.\ large lexical difference, which, for a given pair of sentences, we define as the Jaccard distance between token sets.
We select subsets of the test set where all pairs have a distance greater than $x$ 
and report the accuracy of \BERTSS and \BERTPS, as functions of $x$, in \Figref{fig:perf_mod_size}. 
For reference, we also report the \GPTSS and \GPTPS baselines. 

While \BERTSS and \BERTPS perform significantly better on the subsets with larger modifications, \GPT baselines do not see significant accuracy changes.
The accuracy increase is particularly strong for \BERTPS, probably because pairs with a large lexical difference inherently have more information about their semantic difference.
Despite never seeing sentences in pairs, the accuracy of \BERTSS also increases significantly with modification size. A potential explanation might be that funny sentences for which humans could not find a small modification in order to remove the humor may also be easier to recognize as funny.

\begin{figure}
\centering
\begin{subfigure}[t]{.5\columnwidth}
  \centering
  \includegraphics[width=.93\linewidth]{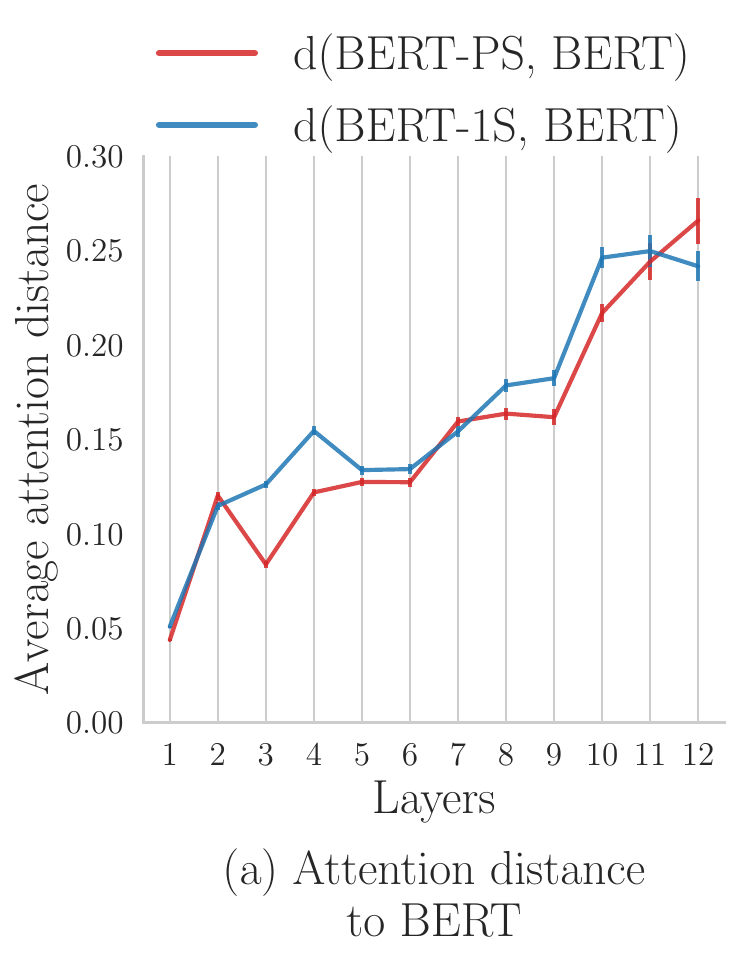}
\end{subfigure}%
\begin{subfigure}[t]{.5\columnwidth}
  \centering
  \includegraphics[width=.97\linewidth]{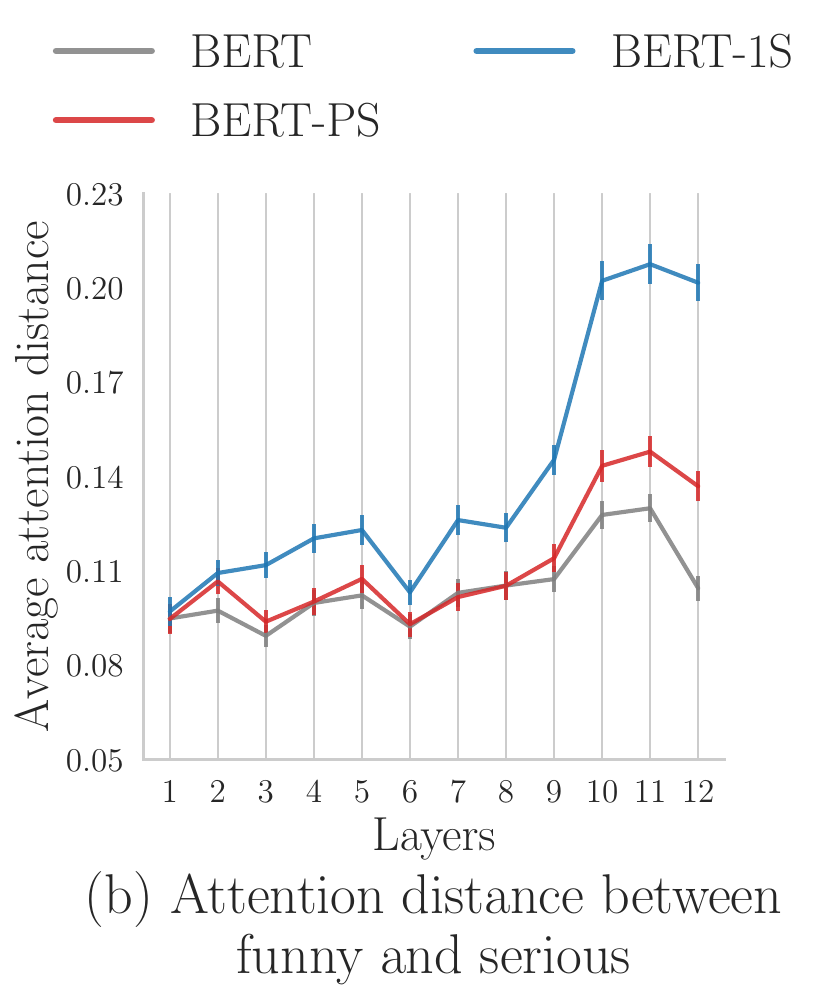}
\end{subfigure}
\caption{Average per-layer attention distance: (a)~between finetuned models (\BERTSS and \BERTPS) and non\hyp finetuned \BERT, for fixed sentences; (b)~between funny and serious sentence in pairs, for fixed models.}
\label{fig:att_per_layers}
\end{figure}

\subsection{Global Attention Patterns}
\label{ssec:attention_pattern}
To better understand \BERTSS and \BERTPS, we now investigate their attention patterns. The models have 12 layers and 12 attention heads per layer, for a total of 144 heads. For a given sentence $s$ of length $|s|$, each head computes $|s|+2$ self\hyp attention distributions (where the $+2$ comes from the two special ``[CLS]'' and ``[SEP]'' tokens).

\xhdr{More finetuning happens in later layers} 
First, we compare the attention distribution patterns of \BERTSS and \BERTPS after finetuning to those of \BERT before finetuning.
We write $A^{M}_{h_i}(s)$ for the $i$-th attention distribution (associated with the $i$-th input token) of model $M$ for head $h$ on sentence $s$. The distance between the attention distributions of two models $M_a$ and $M_b$ at head $h$ on sentence $s$ is given by the average Jensen--Shannon divergence
\begin{equation}
	D^{h}_s(M_a, M_b) = \frac{1}{|s|+2}\sum\limits_{i = 1}^{|s|+2} JS\left(A^{M_a}_{h_i}(s), A^{M_b}_{h_i}(s)\right).
\end{equation}
Following the work of \newcite{clark-etal-2019-bert}, we average this quantity over the test set $\mathcal{T}$ to obtain the average attention distance $D^{h}(M_a, M_b)$ between the two models at head $h$:
\begin{equation}
	D^{h}(M_a, M_b) = \frac{1}{|\mathcal{T}|} \sum\limits_{s \in \mathcal{T}} D^{h}_s(M_a, M_b).
\end{equation}
Furthermore, all heads in the same layer can be averaged to produce an average attention distance $D^{L}(M_a, M_b)$ between the two models at layer $L$.
In \Figref{fig:att_per_layers}(a), we report the attention distance $D^{L}$ between finetuned models (\BERTSS and \BERTPS) and \BERT . For reference, in \Appref{sec:details_att}, we show the attention distances for each head.
For both \BERTPS and \BERTSS, attention patterns significantly drift away from plain \BERT's attention patterns as depth increases. It appears that finetuning modifies \BERT more in later layers, in the more semantic stages, while preserving low-level syntactic processing in the earlier layers.

\xhdr{Attention difference between funny and serious increases with depth}
Next, we measure the difference in attention between a funny sentence and its serious counterpart, computed as follows, for a model $M$, sentence pair $(s_j, s_k)$, and head $h$:
\begin{equation}
    D^{h}_M(s_j, s_k) = \frac{1}{|s_j|+2}\sum\limits_{i = 0}^{|s_j|+2} JS\left(A^{M}_{h_i}(s_j), A^{M}_{h_i}(s_k)\right).
\end{equation}
For this analysis, we have to restrict ourselves to cases where $|s_j| = |s_k|$. 
Again, all heads in the same layer can be averaged to obtain an average attention distance between funny and serious in that layer. Averaging this quantity over all sentence pairs (for \BERT, \BERTSS, and \BERTPS) yields
\Figref{fig:att_per_layers}(b). 
For both \BERTPS and \BERTSS, the difference in attention patterns between funny and serious sentences increases with depth, especially in the last three layers. 
This difference is significantly larger for finetuned models than for \BERT and particularly large for \BERTSS. In fact, for \BERTSS, there is a jump at layer $10$ observed in both \Figref{fig:att_per_layers}(a) and \Figref{fig:att_per_layers}(b); we shall come back to this observation in \Secref{ssec:special_head}, where we study the behavior of one particular head in layer~$10$.

\begin{figure}
    \centering
    \includegraphics[width=0.85\linewidth]{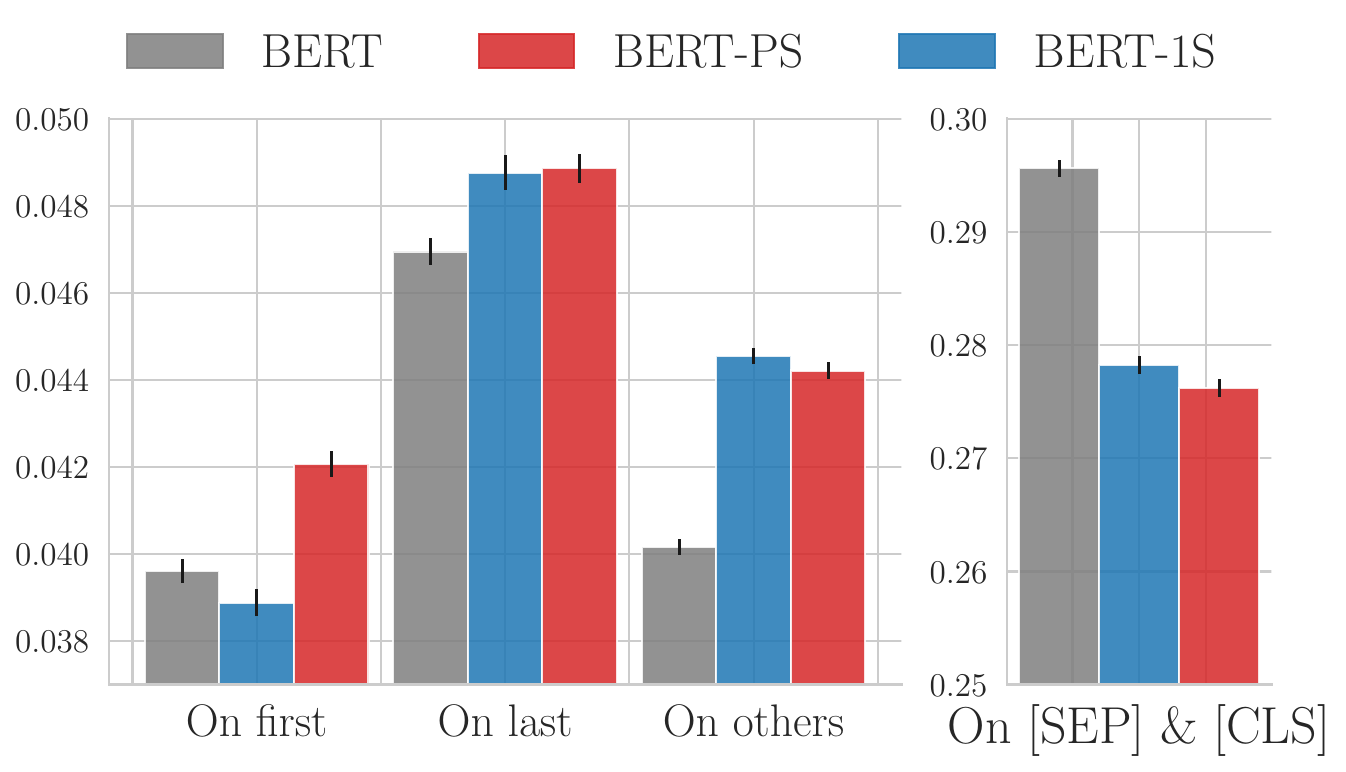}
    \caption{Average total attention paid to special positions and tokens for \BERT, \BERTSS, and \BERTPS.}
    \label{fig:att_pos}
\end{figure}


\xhdr{Boundary tokens are particularly important}
In \Figref{fig:att_pos}, we measure the total amount of attention paid to the first and last words, averaged over the test set, for \BERT, \BERTSS, and \BERTPS. Here, total attention is obtained by summing the attention received by the respective position over all attention distributions in the model.
We observe that, while (non\hyp finetuned) \BERT already attends more to the last word than to other words (as also observed by \newcite{kovaleva-etal-2019-revealing}), both \BERTSS and \BERTPS attend significantly more than \BERT on the last word. Additionally, \BERTPS also attends significantly more to the first word than \BERT. 
Finally, \BERT attends a lot to special tokens, whereas both finetuned models redirect part of this attention toward actual words.
These results confirm prior work, which has established that the humor in satirical headlines tends to be particularly associated with the first word and last word of the headline \cite{west2019reverse,hossain-etal-2019-president}. 




\begin{figure*}
        \centering
        \includegraphics[width=0.85\linewidth]{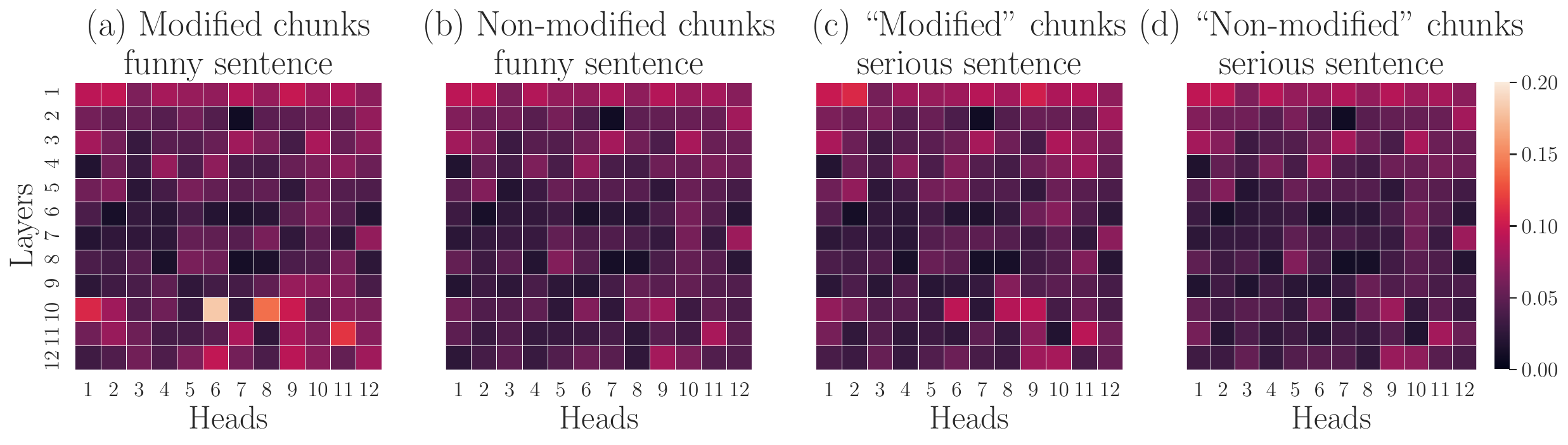}
        \caption{Average attention paid by \BERTSS to
        (a)~the modified (\ie, ``funny'') token in funny sentences,
        (b)~the non\hyp modified tokens in funny sentences,
        (c)~the new token that replaces the ``funny token'' in serious sentences,
        and (d)~the other tokens in serious sentences.
        Lighter colors represent higher average attention.}
        \label{fig:bert_ss_mod}
\label{fig:ss_mod}
\end{figure*}

\subsection{Head 10-6: The ``Laughing Head''}
\label{ssec:special_head}

We found a particularly intriguing attention pattern on head 10-6 of \BERTSS. This head seems to specialize on attending to 
the ``funny token'' of a funny sentence, \ie, the token chosen by a human to remove the humor. (We focus here on pairs where exactly one token from the satirical version was modified.) This is clearly shown by the attention maps of \Figref{fig:ss_mod}, which plot the average attention paid by each head
\begin{enumerate}
\denselist
\item to the modified (\ie, ``funny'') token in funny sentences (\Figref{fig:ss_mod}(a));
\item to the non\hyp modified tokens in funny sentences (\Figref{fig:ss_mod}(b));
\item to the new token that replaces the ``funny token'' in serious sentences (\Figref{fig:ss_mod}(c));
\item to the other tokens in serious sentences (\Figref{fig:ss_mod}(d)).
\end{enumerate}
\Figref{fig:ss_mod}(a) shows that head 10-6 is clearly special and attends strongly to the modified (``funny'') token of funny sentences without activating in other cases. In particular, it does not activate for the same position in serious sentences (\Figref{fig:ss_mod}(c)).
A simple rule that predicts the funny token to be the one to which head 10-6 pays most attention achieves an accuracy of 37\%, nearly five times that of predicting a random token~(8\%).

\newcite{west2019reverse} showed that surface features such as parts-of-speech (POS) tags and position in the sentence were highly associated with whether a token was edited (\eg, final tokens were particularly likely to be edited).
If head 10-6 only recognized such surface features, it should---unlike what we observe empirically---regularly activate for serious sentences as well.
Investigating further, we explicitly compared the ability of head 10-6 to detect the funny token in funny sentences to three baselines:
(1)~predicting the final token,
(3)~predicting the rightmost token with the POS tag overall associated most with edited tokens,
and (3)~predicting the most surprising token (\ie, with the lowest likelihood according to \GPT).
The best baseline (predicting the final token) correctly detects the funny token in $12\%$ of instances, three times less frequently than head 10-6 ($37\%$). Predicting the most frequent POS tag achieves $11\%$, and predicting the most surprising token, $10\%$.
We further tested whether head 10-6 mostly recognizes surprising words by measuring its activation on the ``funny'' position when the respective token is replaced by a random token, finding that in this case head 10-6 activates only $60\%$ as much as with the original token, which indicates that the activation of head 10-6 is only partially due to surprisal.

The existence of such a head is particularly striking given that \BERTSS never  observes pairs of sentences, but only single sentences with a funny or serious label. One could have imagined that, even if a model developed the ability to detect the funny token, this property could be distributed inside the model. Yet, the jumps in attention distances observed in \Figref{fig:att_per_layers}(a) and \ref{fig:att_per_layers}(b) are mostly explained by this one head. 

This supports the hypothesis that \BERTSS learns to detect humor by first identifying a particularly important feature of the data: what is the smallest change that distinguishes funny from serious? We further confirm this insight with a perturbation experiment: for each sentence in the test set, we iteratively mask each word and observe how the classification of \BERTSS changes. Intuitively, we expect that a system recognizing the funny token of a sentence would often switch its decision from funny to serious when the funny token is masked.
In \Tabref{tab:sliding-masking}, we report the percentage of decision changes resulting from masking the modified token \vs\ masking other tokens, for both funny and serious sentences. Masking the modified (funny) token in funny sentences changes the decision from funny to serious $24\%$ of the time, compared to only $13\%$ when other tokens are masked. Furthermore,
which token is masked does not make a significant difference (with respect to decision changes)
in serious sentences. This confirms that the model has, to some extent, learned to recognize not only if a sentence contains humor, but also where the humor is located.

The effect of head 10-6---which we call the ``laughing head''---is large and robust. It remains present and strong even when changing random seeds. Furthermore, we show in \Appref{app:lh_others} that a laughing head also emerges in \dBERT but, interestingly, not in \ROBERTA. 

\begin{table}
        \small
        \centering
        \begin{tabular}{l|cc}
        \toprule

          & Modified token & Other tokens \\
        \midrule
        Funny sentences    & .240 & .134 \\
        Serious sentences & .062 & .095 \\
        \bottomrule
        \end{tabular}
        \caption{Perturbation analysis: Percentage of decisions changed when masking modified vs.\ other tokens. The difference is significant for funny ($p< 10^{-6}$), but not for serious ($p>0.01$), sentences.}\label{tab:sliding-masking}
\end{table}

%% file: 060conclusion.tex
\section{Discussion and Conclusion}
We started by evaluating transformer-based architectures on the task of humor detection
and then leveraged the unique paired structure of the data to obtain novel insights into how transformers deal with the task, finding that
finetuned \BERT models tend to perform better in cases with larger lexical differences between the funny and serious sentences in the pair. 
We also observed varying accuracy across humor types, with models being particularly strong at identifying humor when the funny and serious sentences are opposed along the \emph{non-obscene\slash obscene} dimension, but struggling more with the \emph{good\slash bad intentions} and \emph{non-violence\slash violence} dimensions. 
An analysis of attention patterns revealed that finetuning mostly modifies the last transformer layers and that models attend to funny and serious sentences differently. This difference grows with layer depth significantly more than for non-finetuned \BERT. This indicates that the critical computation takes place in the last transformer layers and that the model picks up on semantic, rather than lexical or syntactic, features. 
We also found that finetuned models redirect part of the attention dedicated to special tokens (``[CLS]'' and ``[SEP]'') by non\hyp finetuned \BERT toward actual words, and particularly towards the last word,
in line with the micro\hyp punchline structure of typical satirical headlines \cite{west2019reverse}.

Our most striking finding pertains to the emergence of a ``laughing head'' that specializes on attending strongly to the funny parts of funny sentences. 
This head alone predicts which words ``contain the humor'' with an accuracy nearly three times as high as the best baseline. 

Our core analyses rely on an investigation of attention heads. \newcite{jain-wallace-2019-attention} warned that attention patterns do not directly imply explanations of model decisions. However, following the subsequent recommendations of \newcite{wiegreffe-pinter-2019-attention}, we always carefully compared the attention of finetuned models against frozen\hyp weight versions (\BERT without finetuning), allowing us to discover significant and meaningful qualitative changes happening during finetuning that enable the model to go from random to significantly\hyp above\hyp random accuracy. Our results thus shed lights on the inner workings of humor detection models.

Overall, this work shows that, although humor detection remains a challenging task,
existing models can already capture highly nontrivial features of what makes satirical headlines funny. Moreover, our characterization of easy vs.\ hard instances can guide future research efforts to further help computational models recognize humor.


%% file: 100appendix.tex
\appendix






\begin{figure}
    \centering
    \includegraphics[width=\linewidth]{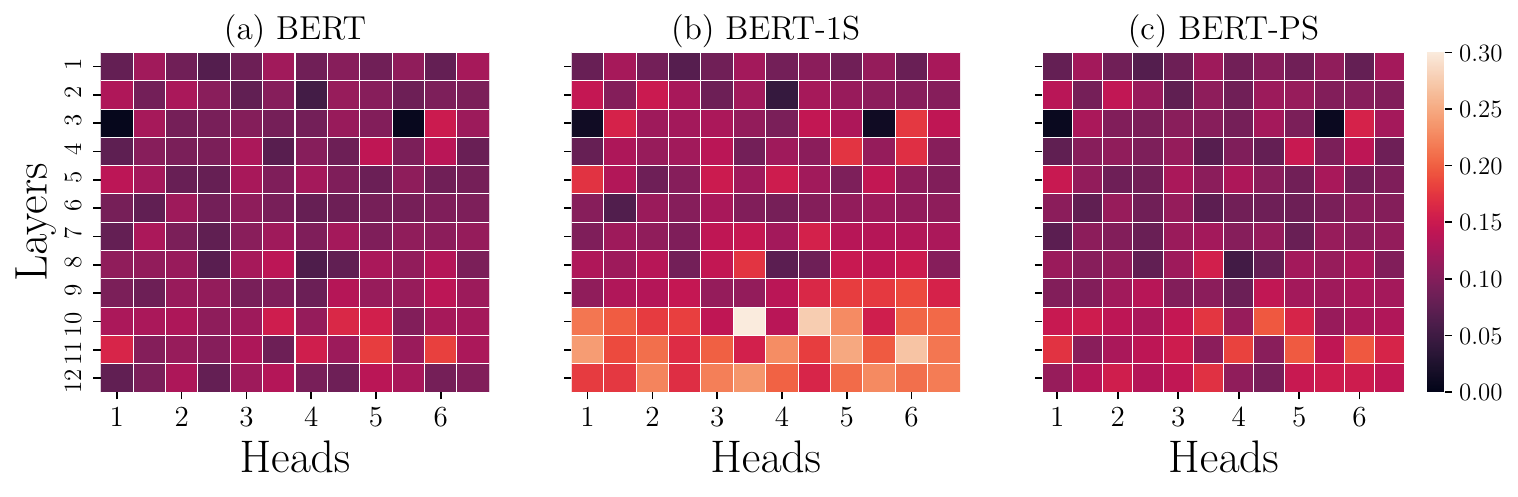}
    \caption{Average, per head, attention distance between funny and serious sentence of each encoder: (a) \BERT, (b) \BERTSS, and (c) \BERTPS}
    \label{fig:funny_serious_ext}
\end{figure}

\begin{figure}
    \centering
    \includegraphics[width=\linewidth]{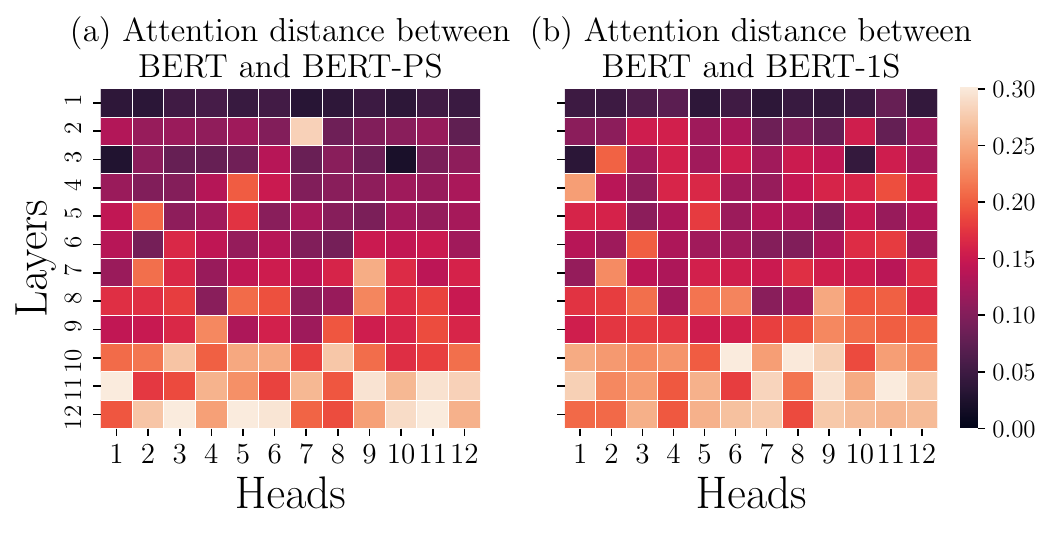}
    \caption{Average, per head, attention distance between finetuned models (\BERTSS in (a) and \BERTPS in (b)) and \BERT}
    \label{fig:att_dist_ext}
\end{figure}

\section{More details on attention patterns}
\label{sec:details_att}
In the main paper, \Figref{fig:att_per_layers}(a) and \Figref{fig:att_per_layers}(b) report attention distances averaged for all heads in a same layer to focus on the effect of depth. For reference, we report in \Figref{fig:att_dist_ext} the same computation as the one resulting from \Figref{fig:att_per_layers}(a) but without averaging heads. Similarly, \Figref{fig:funny_serious_ext} reproduces \Figref{fig:att_per_layers}(b) without averaging heads. With these plots, we can also confirm the conclusions from the main, namely that the finetuned models \BERTSS and \BERTPS differ from the non-finetuned \BERT more in the last layers and the difference between funny and serious is increasing with depth (significantly more in \BERTSS and \BERTPS than \BERT).

\begin{figure*}
    \centering
    \includegraphics[width=\linewidth]{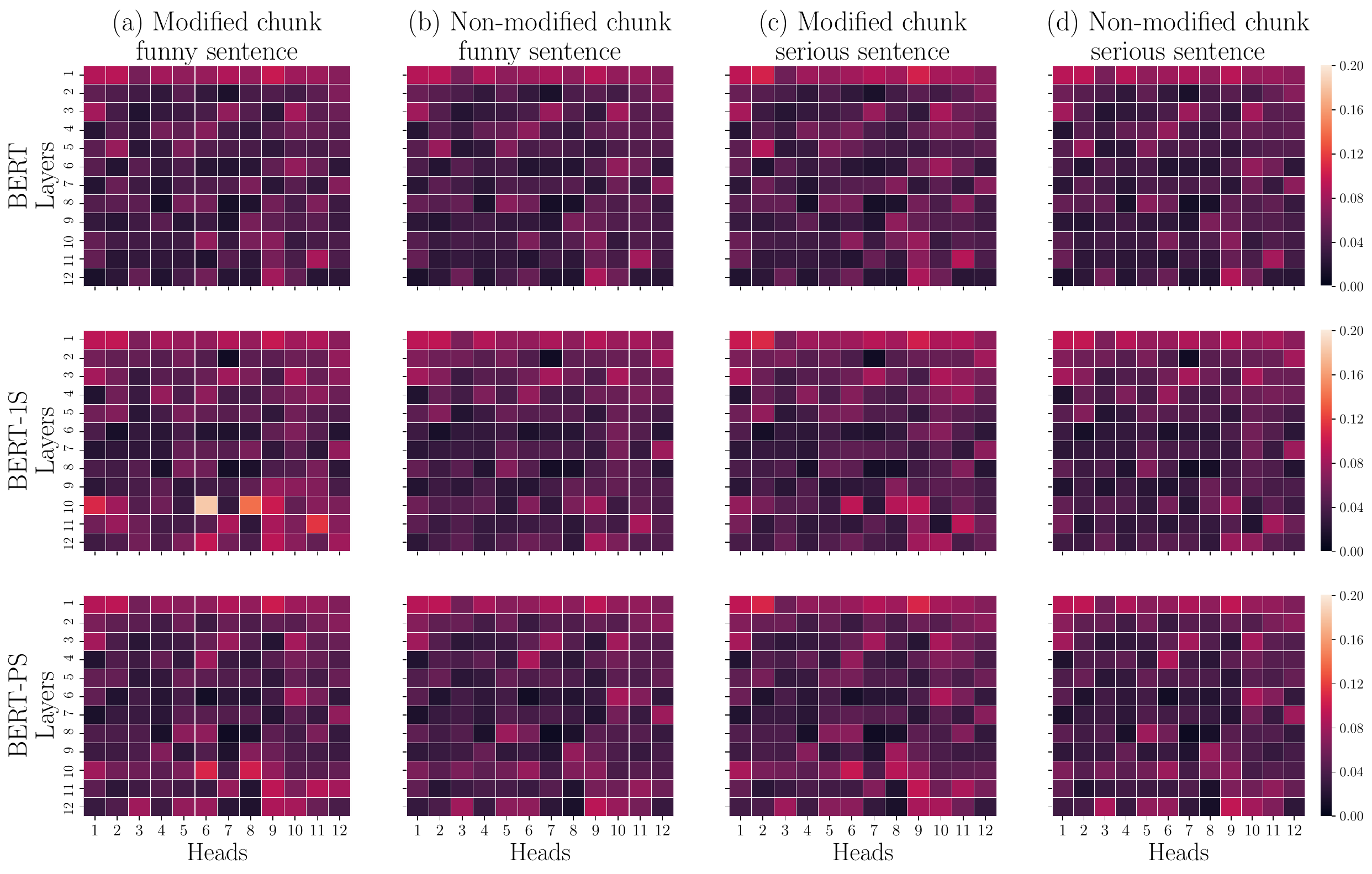}
    \caption{The columns represent attention paid to: (a) modified chunk on funny sentence, (b) non-modified chunk on funny sentences, (c) modified chunk on serious sentences, and (d) non-modified chunk on serious sentences. The first row is \BERT, the second row is \BERTSS (same as \Figref{fig:bert_ss_mod} in the paper), and the last row is \BERTPS. Lighter color represents higher average attention.}
    \label{fig:mod_full}
\end{figure*}

Finally, in the main paper, we identified the special head of \BERTSS by carefully comparing its attention patterns on modified and non-modified chunks for both funny and serious sentences. To confirm that his behavior is really special to \BERTSS, we report in \Figref{fig:mod_full} the same analysis for \BERTPS and \BERT. We indeed observe that this head 10-6 of \BERTSS is special, since, when compared on the same scale, no other head in other models fires that much.


\section{Details about the ``laughing head''}
\label{app:lh_others}

We here provide additional experiments related to the laughing head phenomena observed in the main paper \Secref{ssec:special_head}.

\subsection{The laughing head is where the finetuning happened}
\label{app:lh_base_diff}

It is particularly intriguing to look at the attention distance per head $h$ between funny and serious sentences for \BERTSS, as shown in \Figref{fig:att_dist_ext}~(b). In this figure, the rows are the layers and cells are heads, whose color indicates how large is the average attention difference between funny and serious sentences. We observe that head 10-6 ($6$-th head of layer 10) is particularly different for funny and for serious sentences.

\subsection{The laughing head in other models}

\begin{figure*}
    \centering
    \includegraphics[width=0.80\linewidth]{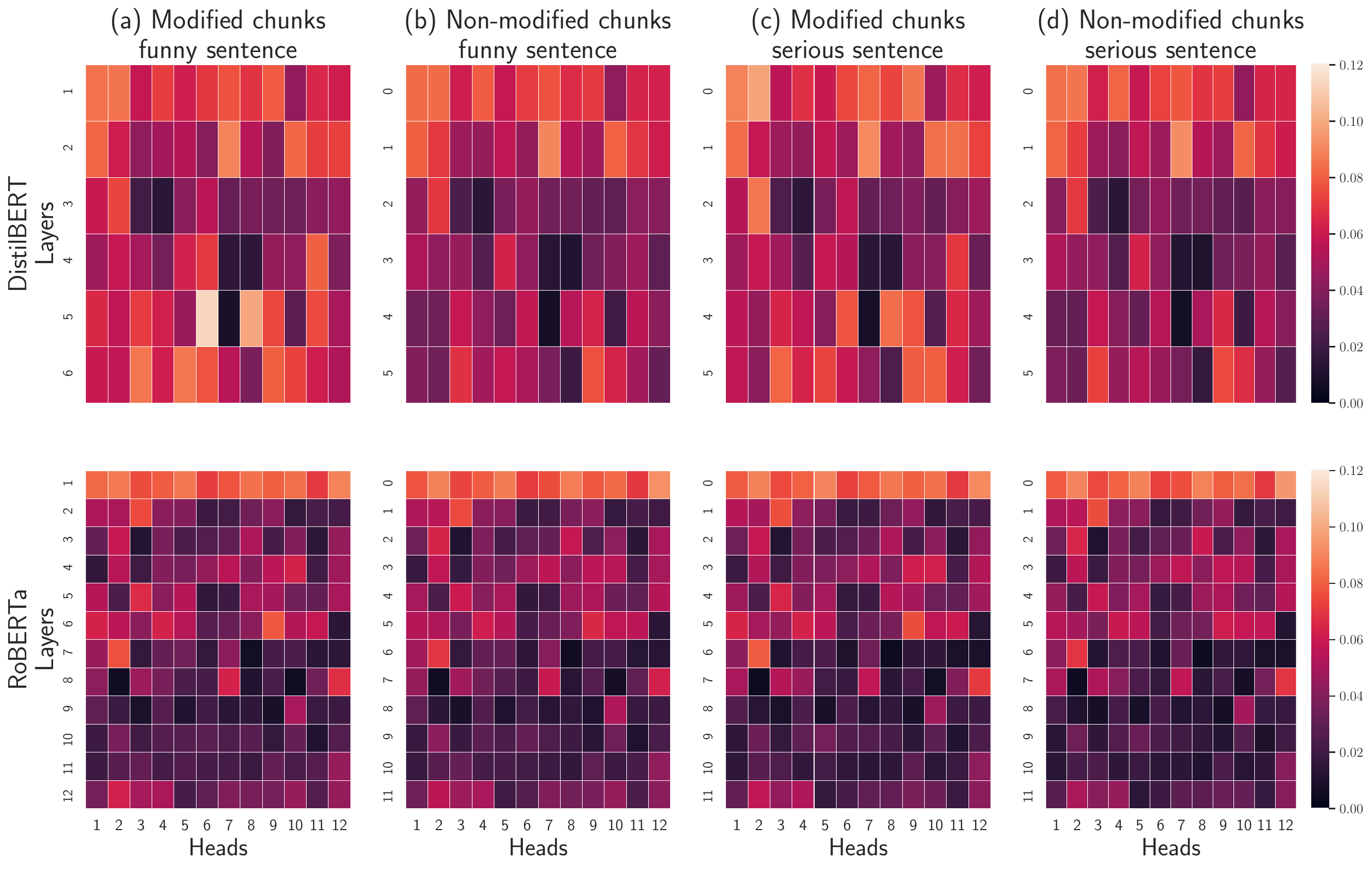}
    \caption{The columns represent attention paid to: (a) modified chunk on funny sentence, (b) non-modified chunk on funny sentences, (c) modified chunk on serious sentences, and (d) non-modified chunk on serious sentences. The first row is \dBERT-1S, the second row is \ROBERTA-1S Lighter color represents higher average attention.}
    \label{fig:other_mod_full}
\end{figure*}

We repeat the experiments described in \Secref{ssec:special_head} with the \dBERT and \ROBERTA architectures also in the single sentence setup. The attention maps on modified/non-modified for funny/serious sentences is reported in \Figref{fig:other_mod_full}.
We see that, to some extent, the head 5-6 in \dBERT also exhibits the same pattern as the head 10-6 of \BERTSS. However, no such head emerges in \ROBERTA.

\subsection{Example of activation of the laughing head}
\begin{figure}
\centering
\begin{subfigure}[t]{.90\columnwidth}
  \centering
  \includegraphics[width=.93\columnwidth]{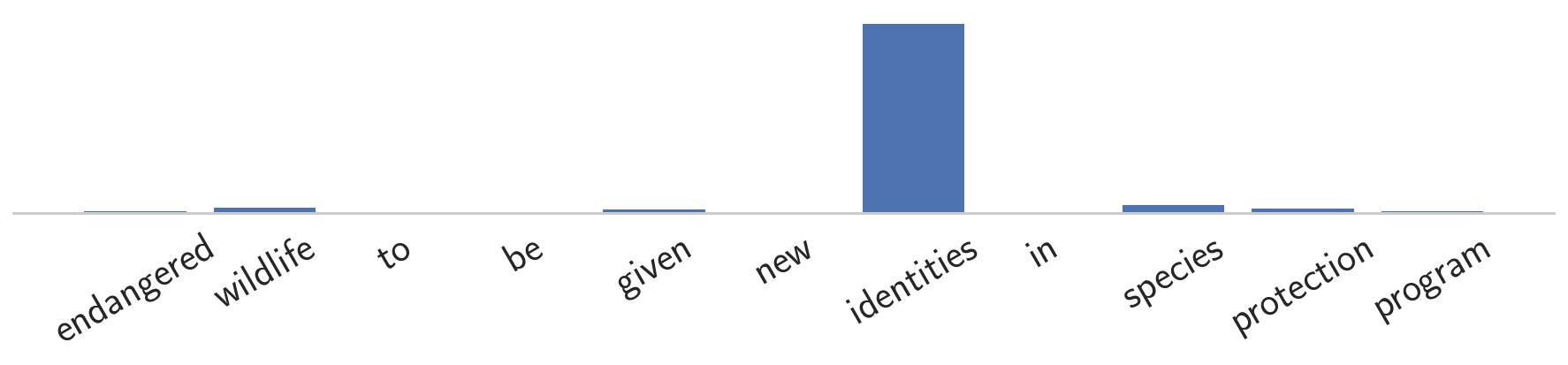}
\end{subfigure}
\begin{subfigure}{.90\columnwidth}
  \centering
  \includegraphics[width=.93\columnwidth]{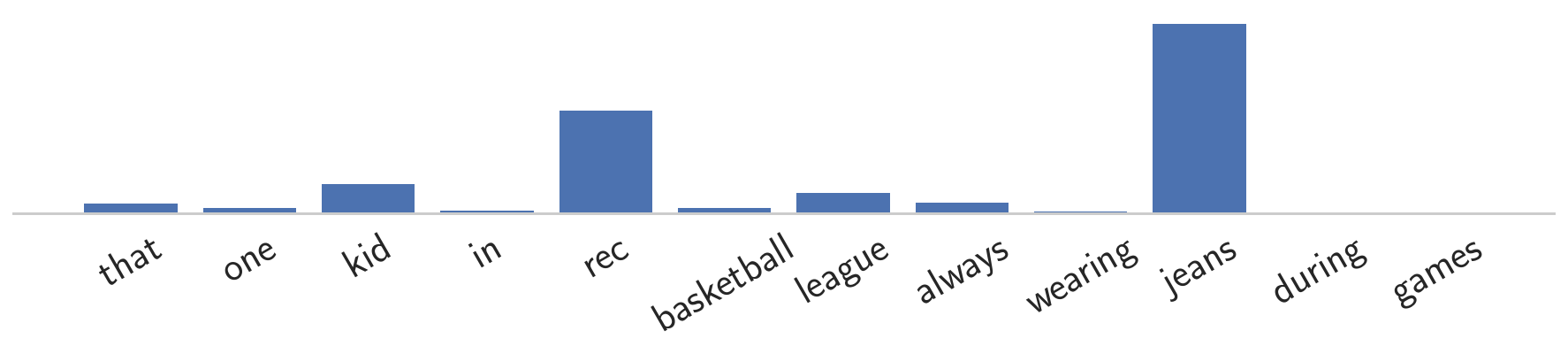}
\end{subfigure}
\begin{subfigure}{.90\columnwidth}
  \centering
  \includegraphics[width=.93\columnwidth]{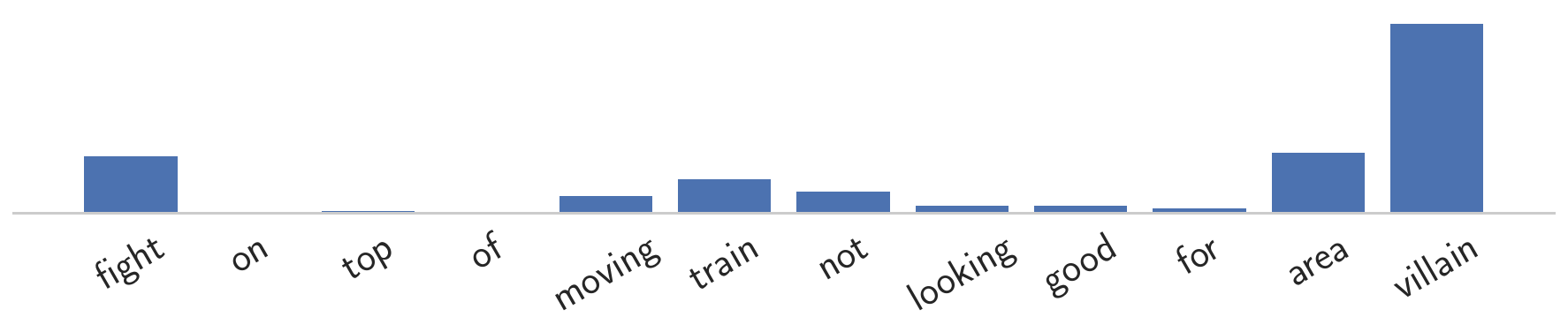}
\end{subfigure}
\begin{subfigure}{.90\columnwidth}
  \centering
  \includegraphics[width=.93\columnwidth]{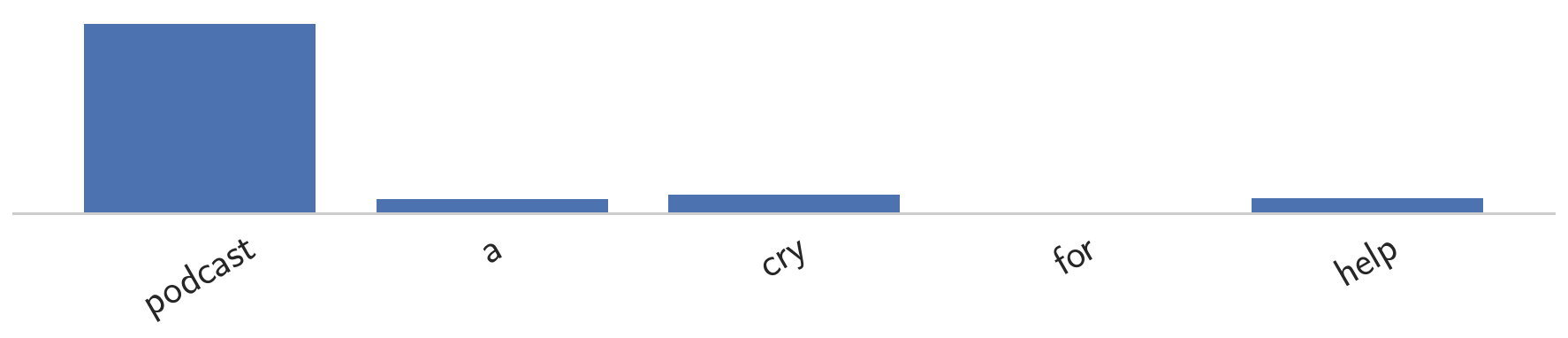}
\end{subfigure}
\begin{subfigure}{.90\columnwidth}
  \centering
  \includegraphics[width=.93\columnwidth]{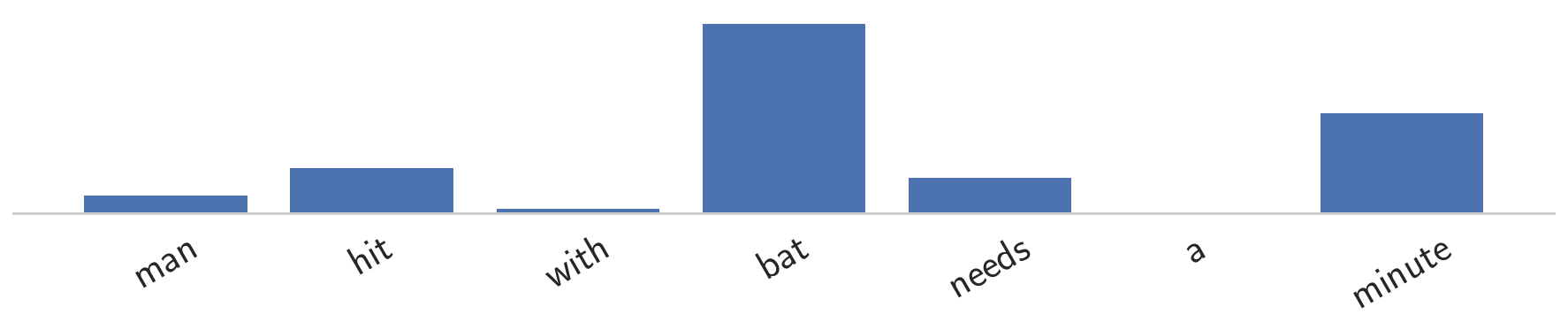}
\end{subfigure}
\begin{subfigure}{.90\columnwidth}
  \centering
  \includegraphics[width=.93\columnwidth]{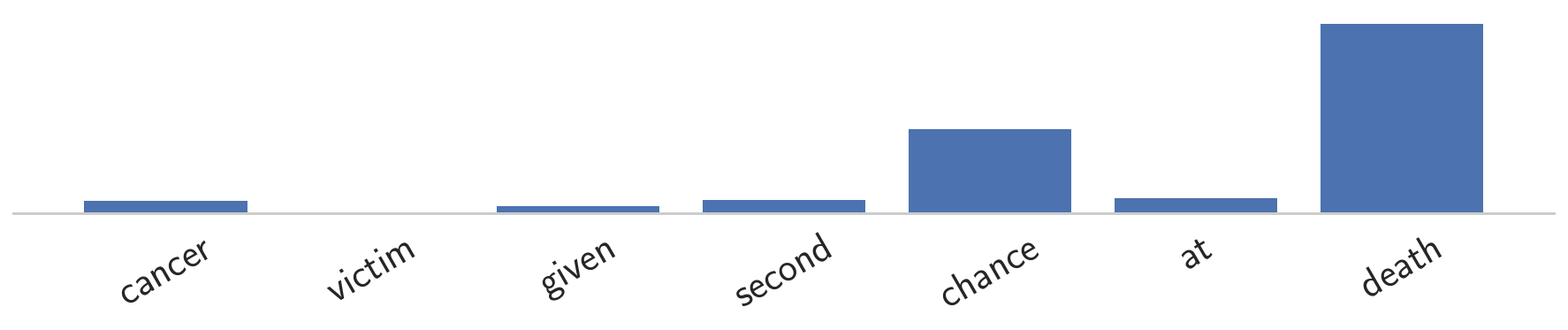}
\end{subfigure}
\begin{subfigure}{.90\columnwidth}
  \centering
  \includegraphics[width=.93\columnwidth]{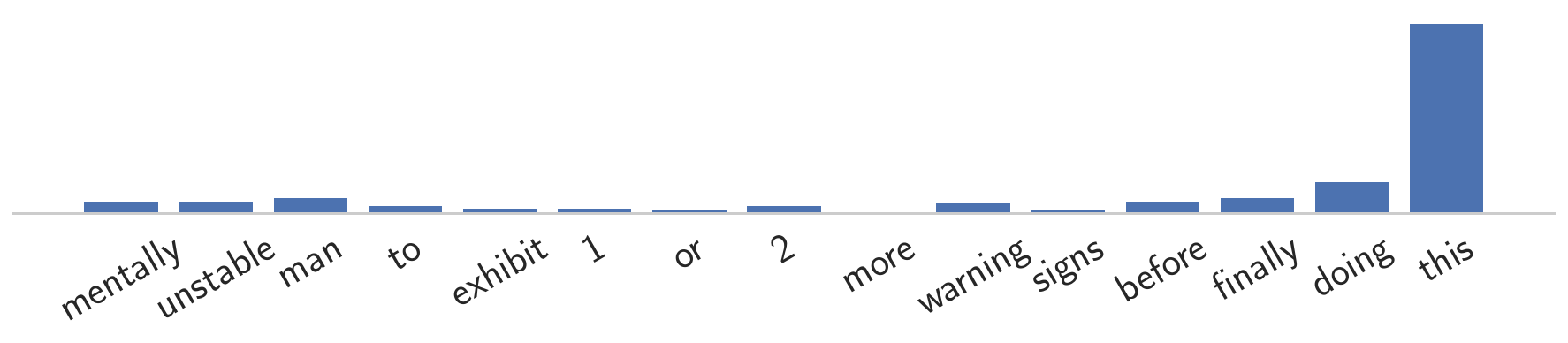}
\end{subfigure}
\begin{subfigure}{.90\columnwidth}
  \centering
  \includegraphics[width=.93\columnwidth]{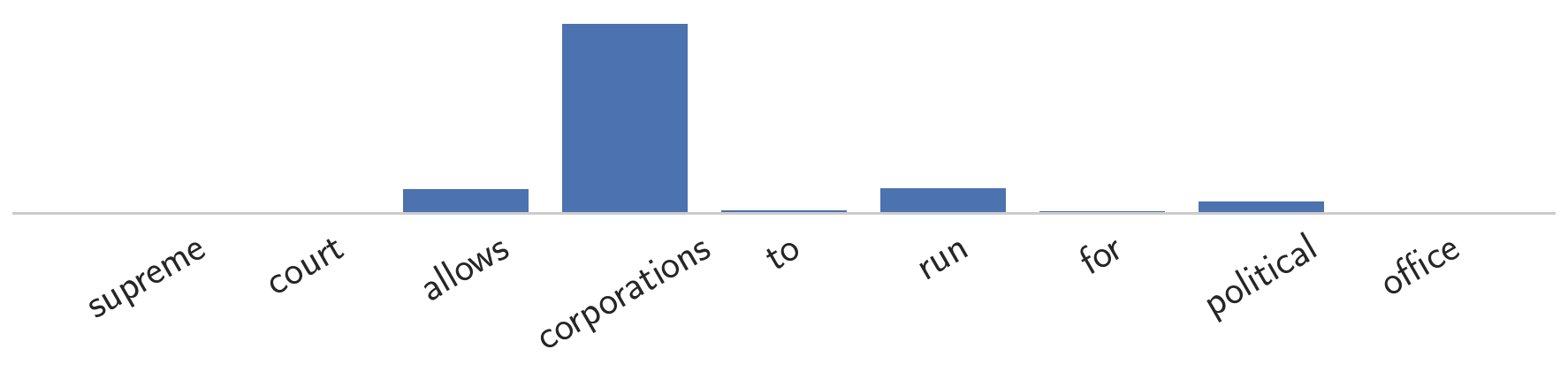}
\end{subfigure}
\begin{subfigure}{.90\columnwidth}
  \centering
  \includegraphics[width=.93\columnwidth]{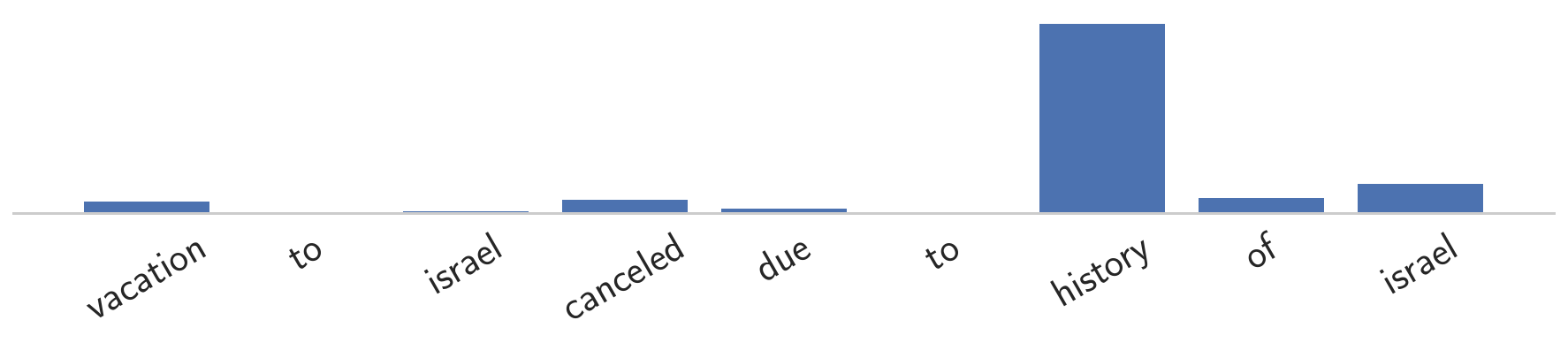}
\end{subfigure}
\begin{subfigure}{.90\columnwidth}
  \centering
  \includegraphics[width=.93\columnwidth]{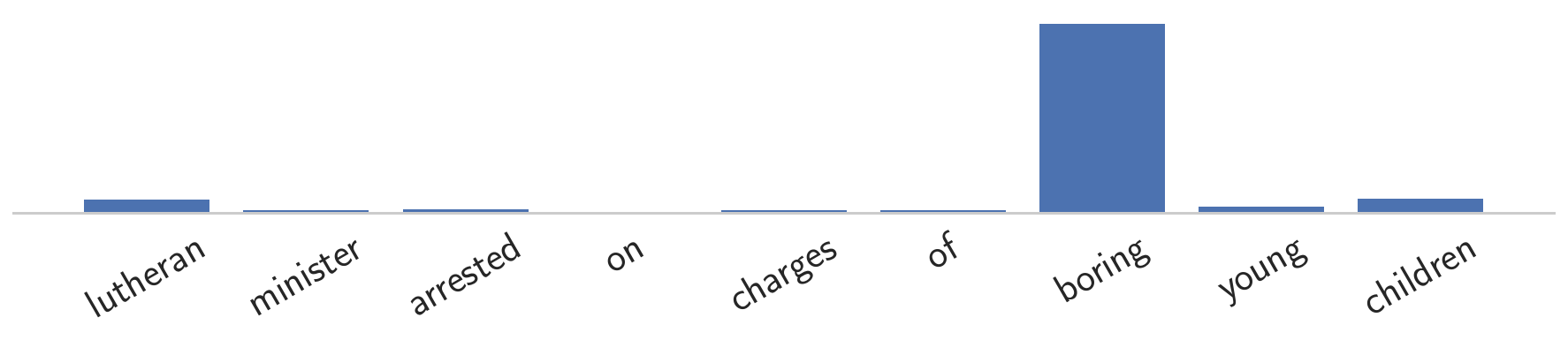}
\end{subfigure}
\begin{subfigure}{.90\columnwidth}
  \centering
  \includegraphics[width=.93\columnwidth]{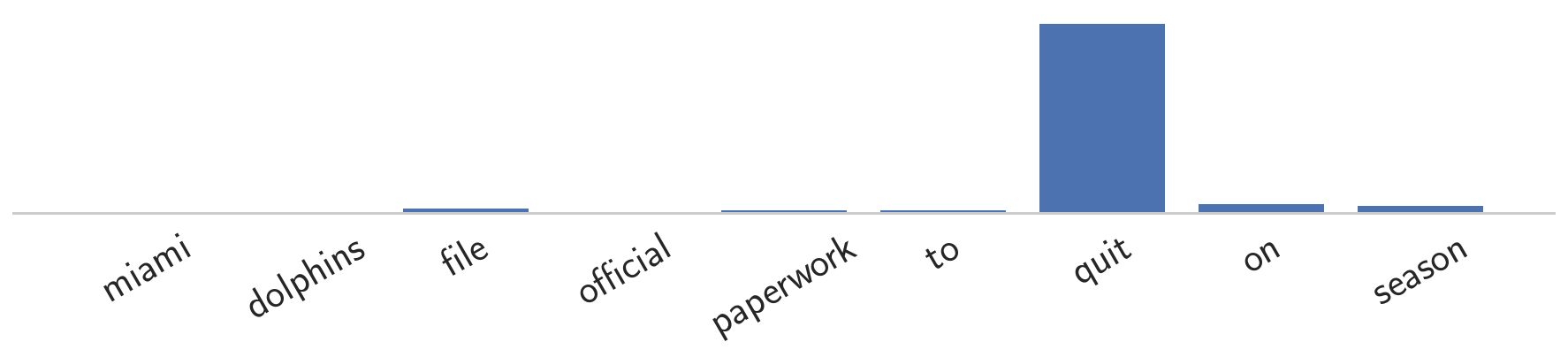}
\end{subfigure}
\caption{Examples of attention distributions on funny sentences by the head 10-6 of \BERT.}
\label{fig:examples}
\end{figure}

For randomly sampled funny sentences from the test set where the laughing head correctly activated on the modified token, we report the total attention paid to each token in the sentence in \Figref{fig:examples}. 

